\documentclass{article}
\usepackage{spconf,amsmath,graphicx}
\usepackage{graphicx}
\usepackage{subfig}
\usepackage{caption}
\usepackage{subfig}
\usepackage{multirow}
\usepackage{float}
\usepackage{acronym}

\newacro{SSL}{Self-supervised Learning} 
\newacro{ECG}{electrocardiagram} 
\newacro{AFib}{Atrial Fibrillation}
\newacro{DEBS}{Distilled Encoding Beyond Similarities}
\newacro{EEG}{electroencephalogram}
\newacro{CPC}{Contrastive Predictive Coding}
\newacro{PCLR}{Patient Contrastive Learning}
\newacro{EMA}{exponential moving average}
\newacro{SBnCL}{Subject-Based non Contrastive Learning}
\newacro{SVC}{ Support Vector Classificatier}
\newacro{BYOL}{Boostrap Your Own Latent}
\newacro{DINO}{Self-Distillation with no Labels}
\newacro{PAR}{Pondered Average Representation}
\newacro{SHHS}{Sleep Heart Health Study}
\newacro{ViT}{Vision Transformer}
\newacro{MLP}{Multilayer Perceptron}
\newacro{MIT-ARR}{MIT-BIH Arrhythmia Database}
\newacro{MIT-AFIB}{MIT-BIH Atrial Fibrillation Database}
\newacro{PCA}{Principal Component Analysis}
\newacro{MAE}{Masked Autoencoders}
\newacro{ML}{Machine Learning}
\newacro{BYOL}{Bootstrap Your Own Latent}
\newacro{ReLU}{rectified linear unit}
\newacro{PSG}{Polysomnography}
\newacro{CINC2017}{Computing in Cardiology Challenge 2017}
\newacro{TF-C}{Time-Frequency Consistency}

\title{Learning Beyond Similarities: Incorporating Dissimilarities between Positive Pairs in Self-Supervised Time Series Learning }
\name{Adrian Atienza, Jakob Bardram, and Sadasivan Puthusserypady}
\address{Department of Health Technology, Technical University of Denmark, Kgs. Lyngby 2800, Denmark}
\begin{document}
%
\maketitle
\begin{abstract}
By identifying similarities between successive inputs, \ac{SSL} methods for time series analysis have demonstrated their effectiveness in encoding the inherent static characteristics of temporal data. However, an exclusive emphasis on similarities might result in representations that overlook the dynamic attributes critical for modeling cardiovascular diseases within a confined subject cohort. 
%
%
%
Introducing \acf{DEBS}, this paper pioneers an \ac{SSL} approach that transcends mere similarities by integrating dissimilarities among positive pairs. The framework is applied to \acf{ECG} signals, leading to a notable enhancement of +10\% in the detection accuracy of \acf{AFib} across diverse subjects. \ac{DEBS} underscores the potential of attaining a more refined representation by encoding the dynamic characteristics of time series data, tapping into dissimilarities during the optimization process. Broadly, the strategy delineated in this study holds the promise of unearthing novel avenues for advancing \ac{SSL} methodologies tailored to temporal data.
\end{abstract}

\section{Introduction}
\label{sec:intro}

\acf{SSL} is a promising approach to machine learning where the model learns to predict certain aspects of a dataset without explicit supervision or labeling of the data. In the context of time series, \ac{SSL} has several benefits: (i) the model learns generic representations that are useful across many tasks, (ii) in some cases, such as physiological signals, the labeling is costly, or the specific task is not known apriori, and (iii) the model learns representations that are more robust across different perturbations incorporated during the recording of the data.

This paper presents a novel \acf{SSL} method for time series analysis, namely the \acf{DEBS}\footnote{Throughout this paper, the term distilled is used in its idiomatic sense rather than in the deep learning sense.}, with a specific focus on the analysis of physiological signals. The underlying concept of this approach is based on the categorization of signal characteristics into two types: (i) inherent static features that account for individual characteristics such as gender and age, and (ii) dynamic features that can reveal transitional states or events experienced by the subjects during the recording, such as heart arrhythmias in \ac{ECG}.
%
%

\begin{figure}[t]
\centering
{\fbox{\includegraphics[width=0.8\linewidth]{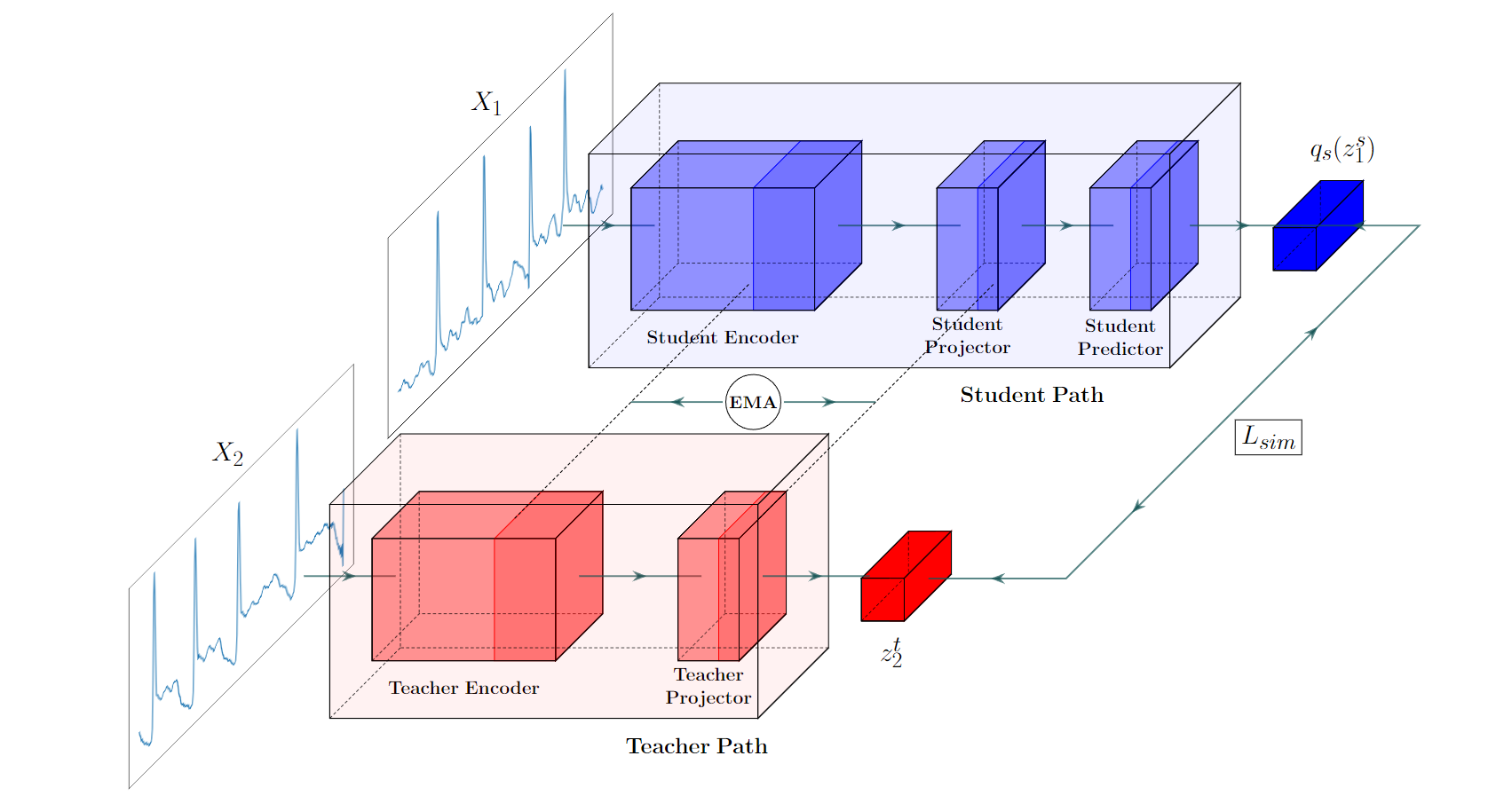}}}
\caption{Similarity path. $X_1$ and $X_2$ belong to the same subject}
\label{fig:sim_path}
\end{figure}

\begin{figure}[t]
\centering
{\fbox{\includegraphics[width=0.8\linewidth]{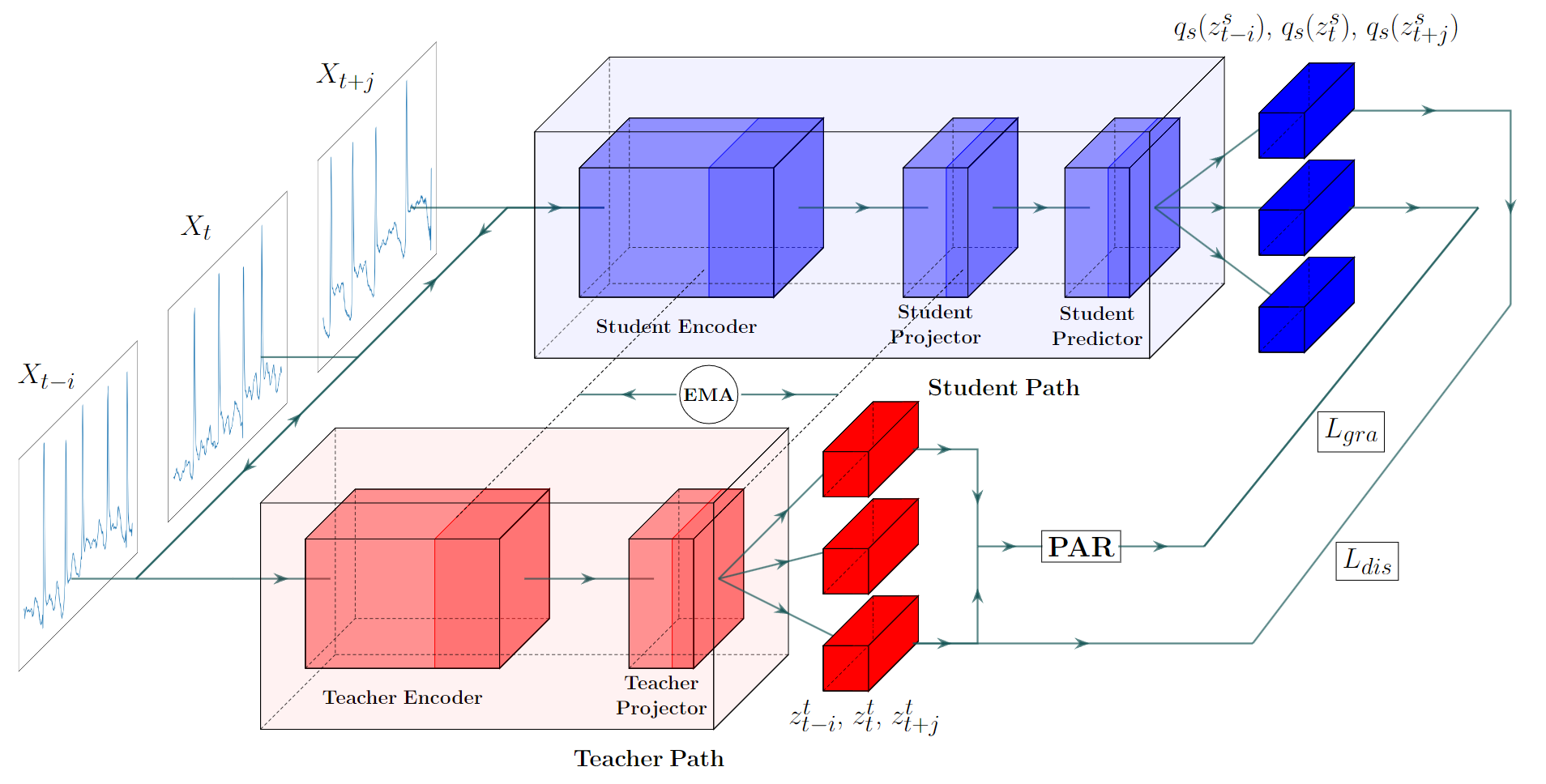}}}
\caption{Dissimilarity path. $X_{t - i}$, $X_{t}$, $X_{t + j}$ belong to the same subject}
\label{fig:dissim_path}
\end{figure}

As such, \ac{DEBS} \textit{not only captures what is common but also the  changes between two positive pairs.} This is achieved by enhancing the dissimilarities between the positive pairs. 
We extend the ongoing research trajectory that considers the naturally obtained multiple views as an organic source of variance in order to avoid data augmentation. In addition, we hypothesize that (i) looking solely for similarities can lead the representations to neglect altogether the variance and, therefore, not to encode meaningful dynamic features contained in the data, and (ii) incorporating a focus on the dissimilarities between positive pairs can result in the representation of the dynamic features and, consequently, an improvement in the performance of downstream tasks such as \ac{AFib} detection. 

In summary, the contributions of this paper are: 
(i)~We introduce \ac{DEBS}, the first \ac{SSL} method that enforces the representations of positive pairs to be dissimilar as a part of the objective function.
(ii)~We show that by incorporating dissimilarities during the optimization process, it increases the \ac{AFib} detection accuracy over 10\%.
(iii)~We open a new approach in which dissimilarities between positive pairs should be considered for learning the dynamic features of the signals when handling temporal data, such as \ac{ECG}.

\section{Related Work}
Different \ac{SSL} methods have recently been designed in the domain of time series analysis, exploiting the characteristics of this kind of data. Mixing-up \cite{mix_up} utilizes the temporal characteristics of the data for a more tailored data augmentation by creating a time series, product of mixing two time series from the same subject. From an alternative point-of-view, the \ac{TF-C} \cite{tfc} produces two variations of time-domain and frequency-domain pairs associated with the input signal. The \ac{SSL} method is developed to identify similarities inherent in these paired representations.

All of the aforementioned studies require the creation of at least one version of the same input in order to train the model to learn an invariant representation with respect to the artificial variance introduced through the use of data augmentation. We consider it a bottleneck in the \ac{SSL} domain, due to (i) data augmentation methods specifically designed for physiological signals are still an ongoing area of development. Recent studies \cite{data_aug_1, data_aug_2} indicate that the optimal approach to data augmentation in this context has yet to be determined. 
(ii) Despite achieving a consensus on effective data augmentation procedures, as seen in the field of Computer Vision, the choice of specific data augmentation techniques remains crucial for the success of the \ac{SSL} method being employed \cite{effect_data_1, effect_data_2}.

In a more promising manner, \ac{PCLR} \cite{PCLR} obviates the need of data augmentation by leveraging the inherent dynamic nature of physiological signals. It avoids using data augmentation by considering the two time series belonging to the same subject as positive pairs.

\section{\acf{DEBS}}\label{sec:method}
%
%
Utilizing this organic multiple views and, thereby obviating the need for data augmentation raises an important implicit question. The object of interest evolves across time while exhibiting changes, i.e., the dynamic characteristics of temporal data. A \ac{SSL} which only considers similarities between these multiple views, will neglect these dynamic characteristics, resulting in a loss of information within the representations and a consequent low performance in identifying events in downstream tasks. Therefore, \textit{learning beyond similarities is essential for capturing the dynamic characteristics of the temporal data.} \ac{DEBS} represents the first \ac{SSL} technique to incorporate dissimilarities between positive pairs as part of the objective in addition to similarities during the training process, with the purpose of driving the representations to reflect what has changed within the signal and therefore, capturing the dynamic characteristics within the representation.

\subsection{Description of DEBS}

    
    


\textbf{Non-Contrastive Method: } The omission of the negative pairs enable the proposed \ac{SSL} methods to surpass the conventional emphasis on similarity. By not considering dissimilarity among negative pairs to avoid mode collapse, dissimilarity can be incorporated among positive pairs. As \ac{BYOL} \cite{byol} framework, \ac{DEBS} incorporates both a teacher network and a student network. While the student network is optimized using Stochastic Gradient Descent (SGD) with respect to the loss function, the teacher network serves as an \acf{EMA} of the student network, effectively operating as a delayed version. This \ac{EMA} updating rule is described below.

\begin{equation}
\xi \leftarrow \tau \cdot \xi+(1-\tau) \cdot \theta,
\end{equation}
where $\tau$, $\xi$, and $\theta$ are the updating hyperparameter, the teacher weights, and the student weights, respectively.

In contrast to the \ac{BYOL} method, \ac{DEBS} integrates two projectors within both the student and teacher networks. Consequently, two predictors are also incorporated into the student network, deviating from using a single predictor. As a result, the encoder generates representations that traverse two distinct paths, namely the similarity path and the dissimilarity path, as termed in this work. The rationale behind this design is to enable the first path to capture static features inherent in the representation while the second path to capture the dynamic features. 

\vspace{0.2cm} \noindent \textbf{Similarity path: } The objective of the student network's predictor is to produce a representation that closely aligns with the one generated by the same path in the teacher network. It is illustrated in Figure \ref{fig:sim_path}. The degree of similarity serves as one of the cost functions in the proposed method, termed the ``Similarity Loss ($\mathcal{L}_{sim}$)'', and it is described as the following:

\begin{equation}
\mathcal{L}_{sim}(\mathbf{z}_2^t, \mathbf{q}_s(\mathbf{z}_1^s)) = 1 - {C}_{Sim}(\mathbf{z}_2^t, \mathbf{q}_s(\mathbf{z}_1^s), 
\end{equation}
where $\mathbf{z}_2^t$ and $\mathbf{q}_s(\mathbf{z}_1^s)$ are the representations for $X_2$ and $X_1$, computed by the teacher and the student network, respectively. $\mathcal{C}_{Sim}$ in Eq.(3) is defined as,

\begin{equation}
\mathcal{C}_{Sim}(\mathbf{x}_1, \mathbf{x}_2)=\frac{x_1 \cdot x_2}{\max \left(\left\|x_1\right\|_2 \cdot\left\|x_2\right\|_2, 1e-8\right)} \text {. }
\end{equation}

\vspace{0.2cm} \noindent \textbf{Dissimilarity path:} In contrast to the similarity path, the dissimilarity path aims to predict representations that exhibit differences between two inputs. To achieve this, the ``Dissimilarity Loss ($\mathcal{L}_{dis}$) (Eq.(4))'' is introduced. It is a cost function specifically designed to guide the optimization process and encourage the model to generate dissimilar representations.  
\begin{equation}
\mathcal{L}_{dis}(\mathbf{z}_{t+j}^t, \mathbf{q}_s(\mathbf{z}_{t-i}^s)) = 1 + {C}_{Sim}(\mathbf{z}_{t+j}^t, \mathbf{q}_s(\mathbf{z}_{t-i}^s)),
\end{equation}
where $\mathbf{z}_{t+j}^t$ and $\mathbf{q}_s(\mathbf{z}_{t-i}^s)$ are the representation vector and the representation prediction for $X_{t+j}$ and $X_{t-i}$, computed by the teacher and the student networks, respectively. \\

\noindent In addition to $\mathcal{L}_{dis}$, we introduce the ``Gradual Loss ($\mathcal{L}_{gra}$)'' as a part of the training objective. We consider that it is not only essential for the representations of two time points drawn from the same subject, $X_{t - i}$ and $X_{t+j}$, to be dissimilar, but also for the
representation of $X_t$ to lie between them. In other words, if a subject’s state evolves from $X_{t - i}$ to
$X_{t+j}$, the representation of $X_t$, i.e., $z_t$, should approximate an intermediate point between these two extremes. This ensures that the temporal evolution is properly captured within the representations. It is described as:
\begin{equation}
\begin{split}
& \mathcal{L}_{gra}(\mathbf{q}_s(\mathbf{z}_{t}^s), \mathcal{PAR}(\mathbf{z}_{t - i}^t, \mathbf{z}_{t + j}^t)) = \\
& 1 - {C}_{Sim}(\mathbf{q}_s(\mathbf{z}_{t}^s), \mathcal{PAR}(\mathbf{z}_{t - i}^t, \mathbf{z}_{t + j}^t)),    
\end{split}
\end{equation}
where \ac{PAR} is the approximation of $\mathbf{z}_{t}$, drawn from $\mathbf{z}_{t-i}$ and $\mathbf{z}_{t+j}$. We do not force $\mathbf{z}_{t}$ to be equally spaced from both $\mathbf{z}_{t-i}$ and $\mathbf{z}_{t+j}$, therefore, we calculate \ac{PAR} as,

\begin{equation}
\mathcal{PAR}(\mathbf{z}_{t - i}^t, \mathbf{z}_{t + j}^t)=\frac{\mathbf{z}_{t - i}^t \cdot j + \mathbf{z}_{t + j}^t \cdot i} {i + j}.
\end{equation}

The overall loss function that is minimized in the dissimilarity path ($\mathcal{L}_{dis path}$) is described as:

\begin{equation}
\begin{split}
& \mathcal{L}_{dis path}(\mathbf{q}_s(\mathbf{z}_{t}^s), \mathbf{q}_s(\mathbf{z}_{t-i}^s), \mathbf{z}_{t - i}^t, \mathbf{z}_{t + j}^t) = \\
& \mathcal{\alpha} \cdot \mathcal{L}_{dis}(\mathbf{z}_{t+j}^t, \mathbf{q}_s(\mathbf{z}_{t-i}^s)) \\
& + \mathcal{L}_{gra}(\mathbf{q}_s(\mathbf{z}_{t}^s), \mathcal{PAR}(\mathbf{z}_{t - i}^t, \mathbf{z}_{t + j}^t)),
\end{split}
\end{equation}
where $\mathcal{\alpha}$ is the dissimilarity coefficient. This dissimilarity path is illustrated in Figure \ref{fig:dissim_path}.

\subsection{Intuitions behind \ac{DEBS}}

\vspace{0.2cm} \noindent \textbf{Intuitions behind a two-step procedure:}
Our belief is that understanding the changes in the input requires first understanding what remains constant. This principle drives \ac{DEBS} to undergo two distinct learning phases. In the initial phase, the method focuses on reducing the variance by ensuring similarity among representations, i.e., encoding the static characteristics of the signals. Once these are adequately captured, \ac{DEBS} drives the model to encode the remaining variance as a source information about the dynamic nature of the data. By doing so, the method does not neglect but understands the remaining variance, ultimately capturing the dynamic characteristics of the temporal data.

\vspace{0.2cm} \noindent \textbf{Intuitions behind the dissimilarity coefficient ($\mathcal{\alpha}$):} Even $\mathcal{L}_{dis}$ is minimized when the representations are entirely dissimilar, it is not what is intended since static features should remain constant throughout the signal. Also the dynamic characteristics need to maintain some level of relational information. Hence, \ac{DEBS} deliberately introduces $\mathcal{\alpha}$ as a regularization factor for lowering the weight of this objective.  

\vspace{0.2cm} \noindent \textbf{Intuitions behind the window size:} An essential consideration in implementing the method is determining the appropriate spacing window size between $\mathbf{X}_{t-i}$ and $\mathbf{X}_{t+j}$, i.e., how much these inputs may be separated in time. This spacing window size must be large enough to accommodate signal changes, yet narrow enough to contain only a single one. If successive changes occur within this window, it can lead to conflicting directions in which these changes are reflected in the representations, thereby $\mathcal{PAR}(\mathbf{z}_{t - i}^t, \mathbf{z}_{t + j}^t)$ may not be aligned with $\mathbf{q}_s(\mathbf{z}_{t}^s)$.

\subsection{Implementation details}

\noindent \textbf{Architecture:}
We use an adaptation of the \ac{ViT}~\cite{vit} model for performing physiological signals. The input data is a time series of 1000 samples, which is split into patches of size 20. The model counts with 6 regular transformer blocks with 4 heads each. The model dimension is set to 128, for a total of 1,192,616 trainable parameters. 

\vspace{0.2cm} \noindent \textbf{\ac{DEBS} implementation:} 
The projectors and predictors are implemented as a two-layer \ac{MLP}, with a dimensionality of 256 and 64, respectively. Batch normalization and \ac{ReLU} operations are incorporated between the two layers of each structure. The \ac{EMA} updating factor ($\tau$), the window size and and the $\mathcal{L}_{diss}$ coefficient are set to 0.995, 2 minutes and 0.1 respectively.

\vspace{0.2cm} \noindent \textbf{\ac{DEBS} optimization:} The model is trained with the \ac{SHHS} dataset~\cite{shhs1, shhs2}. The training procedure consists of 25,000 iterations. After 15,000 iterations, dissimilarities are integrated into the objective function, while similarities are no longer taken into account. Before starting the second step, we update the teacher weights as a copy of the current student weights. We use a batch size of 256, and Adam \cite{adam} with a learning rate of $3e-4$  and a weight decay of $1.5e-6$ as the optimizer. 

\section{Experimental evaluation}
\label{sec:results}
\noindent \textbf{Comparison against SOTA methods:} 
In this experiment, we evaluate \ac{DEBS} against three different baselines: (i) PCLR \cite{PCLR}, (ii) Mixing-Up \cite{mix_up} and (iii) TF-C \cite{tfc}. For this evaluation, a \ac{SVC} \cite{svc} is fitted on top of the representations, based on samples obtained from the \ac{MIT-ARR} database \cite{mit-arr}, and evaluated in two different databases (\ac{MIT-AFIB} \cite{mit-afib} and \ac{CINC2017} \cite{cinc2017}). All used datasets are publicly available in Physionet \cite{physionet}.

We have optimized the same model used in this work, under the same configuration (optimizer, data, batch size and number of iterations), except for the TF-C method, where  their proposed model has been used. This is due to the fact that it requires the use of two encoders instead of one. Note that this model contains approximately 32 million parameters, which is 30x more than our proposed model. To ensure that the model converges, the latter has been optimized over 75K iterations, instead of the 25K iterations proposed in this work. We have saved the model ater 25K, 50K and 75K iterations.

Table \ref{tab:sota_eval} shows that the proposed method clearly outperform all the baselines in the different databases.
%
\begin{table}[ht]
\caption{Comparison agains SOTA SSL Methods}
\label{tab:sota_eval}
\resizebox{\columnwidth}{!}{%
\begin{tabular}{l|cccc}
\hline
\multicolumn{1}{c|}{Dataset}                                  & SSL Method    & Accuracy (\%) & Sensitivity  (\%) & Specificity (\%) \\ \hline
\multirow{6}{*}{MIT AFIB}               & PCLR \cite{PCLR}         & 72.7          & 65.6              & 78.9             \\
                                                              & Mixing-Up \cite{mix_up}     & 65.0          & 60.5              & 67.2             \\
                                                              & TF-C (25K) \cite{tfc}  & 72.2          & 65.0              & 78.5             \\
                                                              & TF-C (50K) \cite{tfc} & 69.8          & 62.3              & 76.2             \\
                                                              & TF\_C (75K
                                                              ) \cite{tfc}& 71.3          & 65.9              & 76.4             \\
                                                              & \textbf{DEBS} & \textbf{77.5} & \textbf{75.6}     & \textbf{79.5}    \\ \hline
\multirow{6}{*}{\ac{CINC2017} (Training)}   & PCLR\cite{PCLR}          & 63.5          & 21.2              & 93.0             \\
                                                              & Mixing-Up\cite{mix_up}     & 68.0          & 20.5              & 90.9             \\
                                                              & TF-C (25K)\cite{tfc}    & 62.4          & 20.5              & 92.8             \\
                                                              & TF-C (50K)\cite{tfc}    & 62.0          & 20.3              & 92.7             \\
                                                              & TF-C (75K)\cite{tfc}    & 62.4          & 20.6              & 92.9             \\
                                                              & \textbf{DEBS}          & \textbf{78.2} & \textbf{34.0}     & \textbf{95.3}    \\ \hline
\multirow{6}{*}{\ac{CINC2017} (Validation)} & PCLR \cite{PCLR}          & 70.6          & 43.3              & 89.2             \\
                                                              & Mixing-Up \cite{mix_up}     & 66.0          & 36.0              & 82.8             \\
                                                              & TF-C (25K) \cite{tfc}    & 69.2          & 42.0              & 42.0             \\
                                                              & TF-C (50K) \cite{tfc}   & 65.5          & 38.0              & 87.0             \\
                                                              & TF-C (75K) \cite{tfc}   & 67.7          & 40.7              & 89.9             \\
                                                              & \textbf{DEBS}          & \textbf{81.8} & \textbf{59.0}     & \textbf{92.7} \\ \hline            
\end{tabular}
}
\end{table}

\vspace{0.2cm}

\begin{figure}
\centering
\subfloat[AFib classification accuracy over iterations]
{\fbox{\includegraphics[width=0.42\columnwidth]{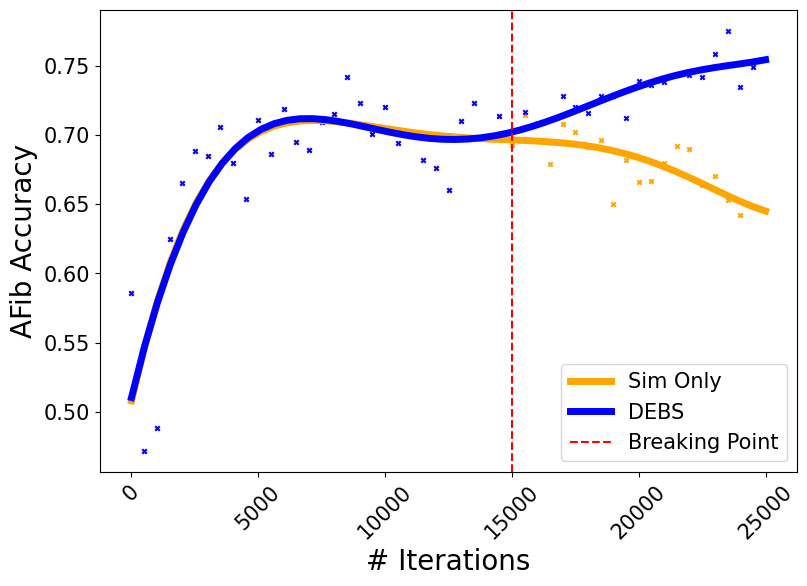}\label{fig:figure1}}}
\qquad
\subfloat[Effect of incorporating dissimilarities]{\fbox{\includegraphics[width=0.42\columnwidth]{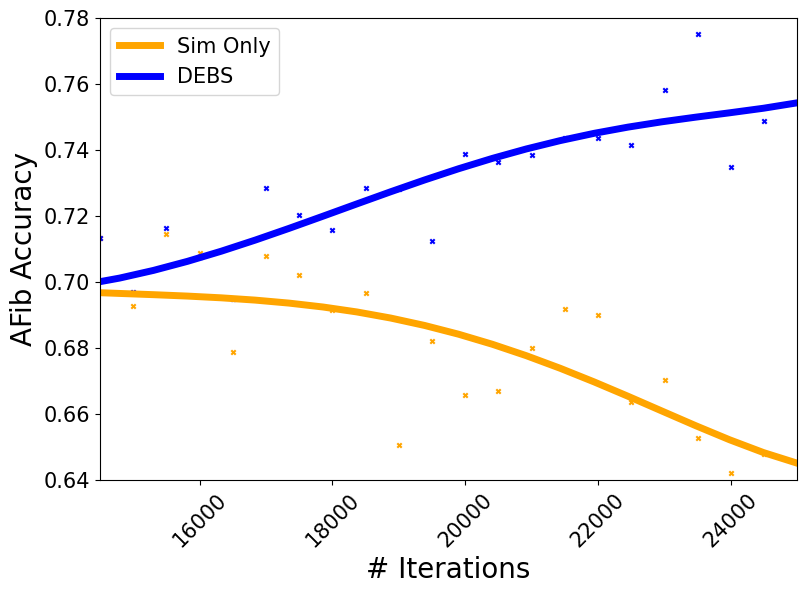}\label{fig:figure2}}}
\caption{Differences between incorporating and not incorporating dissimilarities}
\label{fig:trainingprocess}
\end{figure}

\noindent \textbf{The effect of incorporating dissimilarities:} 
To comprehend the impact of integrating the dissimilarity path, an analysis is conducted on the performance of the \ac{MIT-ARR} $\rightarrow$ \ac{MIT-AFIB} framework during the optimization process, with evaluations performed every 500 iterations. The results, illustrated in Figure \ref{fig:trainingprocess}, highlight that exclusive emphasis on similarities leads to a degradation in model performance. Conversely, the incorporation of dissimilarities contributes to a consistent enhancement in model performance, which leads to a difference of + 10\% at the end of the training procedure.

\begin{figure}
\centering
\subfloat[Subject-based PCA]
{\fbox{\includegraphics[width=0.532\columnwidth]{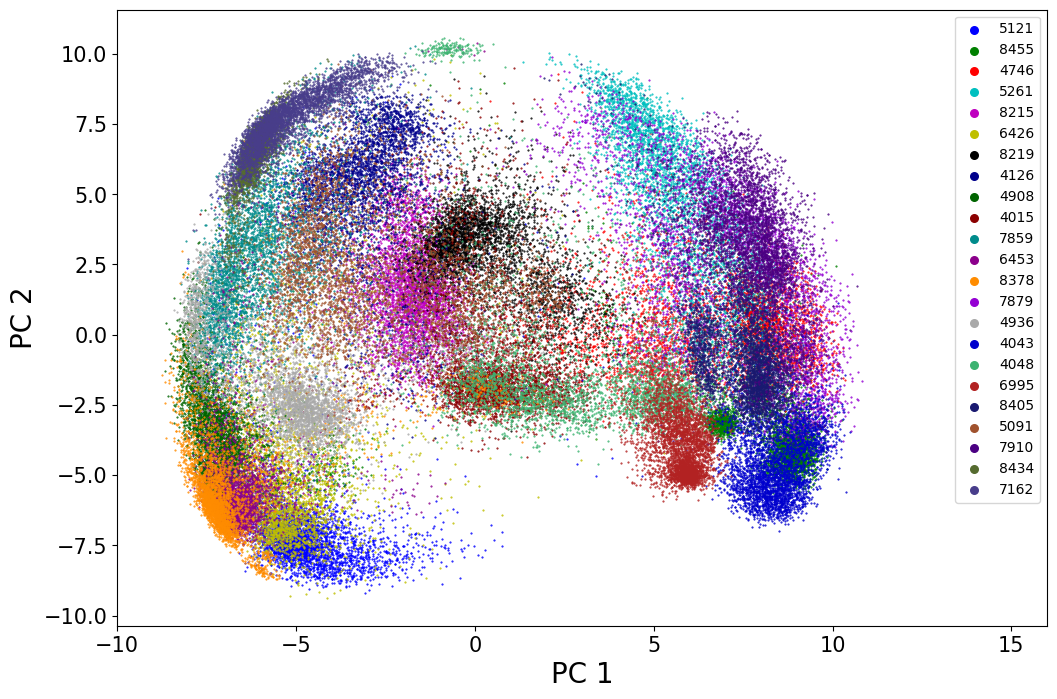}\label{fig:subject_pca}}}
\quad
\subfloat[Event-based PCA]{\fbox{\includegraphics[width=0.368\columnwidth]{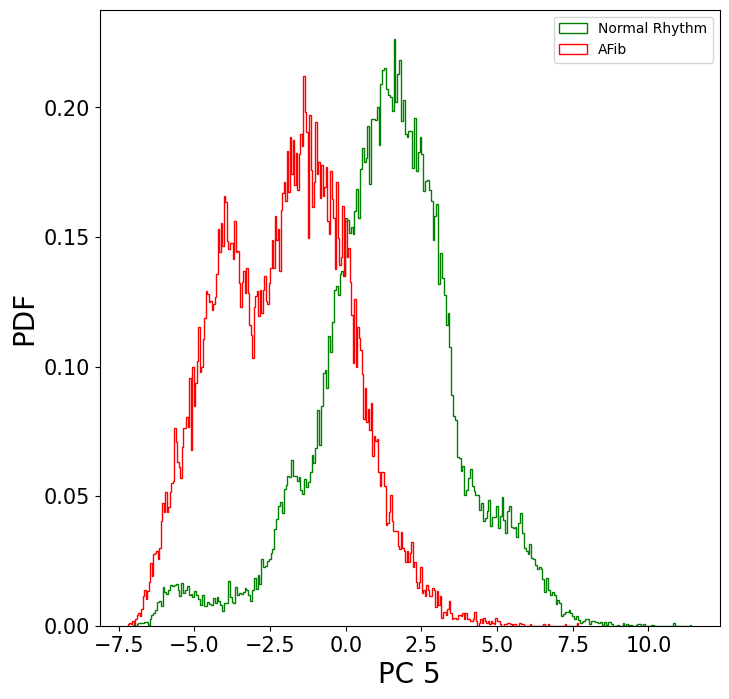}\label{fig:afib_pca}}}
\caption{PCA Analysis on \ac{MIT-AFIB} representations}
\label{fig:pca}
\end{figure}

\vspace{0.2cm}\noindent \textbf{Study on learned representations} The fundamental concept underpinning this approach revolves around the integration of dissimilarities to facilitate the encoding of both static and dynamic features within the model. To ascertain the validity of this premise, a Principal Component Analysis (PCA) \cite{pca} is executed on the generated representations from \ac{MIT-AFIB} database. As depicted in Figure \ref{fig:subject_pca}, temporal segments originating from the same individual exhibit analogous positions in Principal Components 1 and 2, irrespective of the individual's status. Conversely, Principal Component 5 yields distinct values between AFib and normal rhythm (NR) classifications, regardless of the individual, as shown in Figure \ref{fig:afib_pca}.

\noindent \textbf{Discussion of the results:}
This study has demonstrated that by incorporating dissimilarities during the training process, we can enable both static and dynamic characteristics to be captured within the representation, as they are projected separately into distinct components. It leads to a significant difference in the proposed downstream task at the end of the training procedure. 
These results shows substantial support for hypothesis: (i) looking solely for similarities can lead the representations to neglect meaningful dynamic features contained in the data, and (ii) incorporating the dissimilarities between positive pairs results in the representation of the dynamic features and, consequently, an improvement in the performance of downstream tasks such as \ac{AFib} identification. 


%



\section{Conclusion}
In this paper, we have presented \ac{DEBS}, the first \ac{SSL} method that incorporates dissimilarities between positive pairs. We have shown that by incorporating this new objective function, the representations capture not only the static nature of the data but also the dynamic features. It leads to a significant improvement of over 10\% when evaluated on \ac{AFib} classification in dynamic \ac{ECG} time series data. More generally, the approach presented in this paper may open new perspectives for improved \ac{SSL} methods for handling temporal data. 

\vfill\pagebreak
%
\bibliographystyle{IEEEbib}
\bibliography{strings,refs}

\end{document}